\documentclass[runningheads]{llncs}
\usepackage[T1]{fontenc}
\usepackage{graphicx}
\usepackage{svg}
\usepackage{wrapfig}
\usepackage{float}
\begin{document}
\title{Practical Handling of Dynamic Environments in Decentralised Multi-Robot Patrol}
\titlerunning{Patrol in dynamic environments}
%
\author{James C. Ward$^1$\orcidID{0000-0002-8421-6372} \and
Arthur Richards$^{1,2}$\orcidID{0000-0001-9500-5514} \and
Edmund R. Hunt$^1$\orcidID{0000-0002-9647-124X}}
\authorrunning{J. Ward et al.}
%
\institute{$^1$University of Bristol, Bristol, United Kingdom \\ $^2$Bristol Robotics Laboratory, Bristol, United Kingdom \\
\email{\{james.c.ward,edmund.hunt\}@bristol.ac.uk}}
\maketitle              
\begin{abstract}

Persistent monitoring using robot teams is of interest in fields such as security, environmental monitoring, and disaster recovery. Performing such monitoring in a fully on-line decentralised fashion has significant potential advantages for robustness, adaptability, and scalability of monitoring solutions, including, in principle, the capacity to effectively adapt in real-time to a changing environment. We examine this through the lens of multi-robot patrol, in which teams of patrol robots must persistently minimise time between visits to points of interest, within environments where traversability of routes is highly dynamic. These dynamics must be observed by patrol agents and accounted for in a fully decentralised on-line manner. In this work, we present a new method of monitoring and adjusting for environment dynamics in a decentralised multi-robot patrol team. We demonstrate that our method significantly outperforms realistic baselines in highly dynamic scenarios, and also investigate dynamic scenarios in which explicitly accounting for environment dynamics may be unnecessary or impractical.

\end{abstract}
\section{Introduction}

Effective deployment of multi-robot teams to hazardous or communication denied environments is of obvious applicability to a range of fields including security~\cite{michal:security}, environmental monitoring~\cite{dunbabin:monitoring}, and disaster recovery~\cite{ghassemi:disaster}. One desirable behaviour in these scenarios is persistent monitoring --- the long-term repeated observation of points of interest in an environment. This is especially relevant in scenarios where human presence may be undesirable due to the presence of environmental hazards~\cite{bird:robots_for_patrol}, and those in which traditional static surveillance systems may be impractical due to the absence of power or communication infrastructure. It is possible that the traversability of such environments varies significantly over the course of a monitoring scenario --- in these cases, handling unknown environmental dynamics is of vital importance to the effective deployment of a multi-robot team. Here, we introduce a method to adapt patrol behaviour to time-varying route traversability for arbitrary patrol strategies.

\subsection{The Multi-Robot Patrolling Problem}

The Multi-Robot Patrolling Problem (MRPP) is a popular formalisation of this problem, and has seen considerable attention in the literature~\cite{huang:survey}\cite{basilico:survey}. The goal of this problem is typically framed as the persistent minimisation of maximum or average idleness~\cite{machado:idleness_paper}, defined as the time between successive visits to a point of interest in an environment, by a team of mobile autonomous robots. Points of interest in the environment are used to define a patrol graph, where the vertices of the graph represent points to be frequently visited, the edges represent traversable routes between these points, and the edge weights correspond to the time taken for an agent to traverse an edge. A common variant of this problem is ``adversarial'' patrol~\cite{huang:survey}, where rather than idleness minimisation, the goal of the patrol team is to prevent a hostile agent from gaining undetected access to the environment, but in this work we only consider idleness-minimising patrol.

\subsection{Centralised versus decentralised patrol}

In the single-agent case, the patrol tour that persistently minimises maximum idleness on a patrol graph can be found by solving the Travelling Salesman Problem (TSP) on the patrol graph~\cite{chevaleyre:tsp}. Expanding this to multiple agents adds considerable complexity --- approaches exist that attempt to solve the $n$-agent case by partitioning the patrol graph into $m \leq n$ subgraphs and assigning one or more agents to follow a TSP tour on each subgraph~\cite{afshani:older}\cite{afshani:newer}, at considerable computational cost and with large approximation factors. If $m = n$, this is known as the ``min-max vertex cycle cover problem''~\cite{yu:mmccp}, and has been examined considerably in the literature, including within the context of multi-robot patrol~\cite{scherer:mmccp}. Regardless of their proximity to optimality, any attempts to pre-compute optimal cyclic agent trajectories have significant downsides. Any agent failures or changes to the environment have the potential to invalidate a pre-planned solution, and adding new agents to improve performance would require recalculating all trajectories, possibly at significant computational cost.

Consequently, fully on-line approaches, based on real-time, short-horizon decision making are seen as practical approaches and are popular in the literature. Many on-line approaches have the additional advantage of being easily decentralisable, which is potentially advantageous~\cite{durfee:decentralisation} in communication denied environments (i.e. environments in which long range communication may be impossible or only intermittently possible) due to removing any reliance on communication with a central controller. A large number of high performing on-line strategies have been previously published~\cite{portugal:sebs_paper}\cite{portugal:cbls_paper}\cite{yan:er_paper}\cite{ward:suns_mns}\cite{farinelli:dta_paper}, based on a range of heuristic approaches to idleness minimisation and inter-agent coordination. Typically these strategies operate with short-horizon online decision making, whereby upon arriving at a vertex of the patrol graph, they select a neighbouring vertex to next travel to based on edge weights and vertex idlenesses of the patrol graph and any information or intentions communicated by other agents. In this way, decentralised online patrol strategies treat the patrol problem as a POMDP. However, while potential adaptability to dynamic environments is a key benefit of these solutions, little work exists that tackles it explicitly. Examination has been made of modifying some existing strategies to adapt to random edge removal on the patrol graph~\cite{dynamic:2017}, and ``environment dynamics'' within the context of an environment containing dynamically varying reward has been considered~\cite{confusing:2019}. However, to our knowledge no attempts have been made to introduce robustness to general time-varying environments to arbitrary patrol methodologies.

\subsection{Our Contributions}

We address this gap in the literature by presenting a practical approach for handling environments with time-varying traversability in multi-robot patrolling. Our proposed methods are supplemental to existing on-line decentralised patrol algorithms, allowing for existing patrol strategies to be used without modification in the presence of dynamic environments. We also present examination of scenarios in which explicitly accounting for environment dynamics may or may not be necessary --- in some cases, heavily time-varying environments can be treated as if they were static without loss of performance, as we discuss later.

\section{Problem definition}

\subsection{Idleness-Minimising Patrol}

We model the environment to be monitored as a patrol graph $\mathcal{G}$, comprising a set of vertices $\mathcal{V}$ connected by edges $\mathcal{E}$ with weights $\mathcal{W}$. $\mathcal{V}$ corresponds to points of interest in the environment to be frequently visited, $\mathcal{E}$ corresponds to traversable routes between these points, and $\mathcal{W}$ corresponds to the travel times along these routes. $\mathcal{G}$ may be considered undirected under the assumption that routes have the same travel times in each direction, but may also be considered directed when this assumption does not hold, for example in the case of road networks, as we examine in Section~\ref{section:real-world}. The idleness $I_{v, t}$ of a vertex $v$ at time $t$ is defined as the time in seconds since that vertex was most recently visited by a patrol agent. Mean graph idleness $\overline{I}_\mathcal{G}$ is defined as $\frac{1}{T|\mathcal{V}|}\sum_{t=0}^T\sum_{v=0}^{|\mathcal{V}|}I_{v,t}$, i.e. the mean idlenesses of all vertices over the entire scenario time duration. Maximum graph idleness, denoted $\max I_{\mathcal{G}}$, is the maximum idleness over all vertices over the entire scenario time duration. A patrol team of $n$ agents employs a fixed patrol strategy with the intention of minimising either $\overline{I}_\mathcal{G}$ or $\max I_{\mathcal{G}}$.

\subsection{Environment Dynamics}

We model environment dynamics as time-dependent variation of $\mathcal{W}$, potentially around some assumed baseline value, reflecting changes in the time taken to traverse an edge. These variations may result in agents taking longer to traverse an edge, for example when modelling congestion or obstructions, or may result in agents taking less time to traverse an edge, for example in cases where agent speed is influenced by wind speed and direction. The baseline value may not be known accurately by the patrol team, or may reflect some prior belief about the environment --- in this way, the handling of \textit{dynamic} environments can be considered adjacent to the handling of \textit{uncertain} environments. We consider that an agent can only truly know the weight of an edge at any given time by traversing it --- while it may not be generally possible to predict the travel time of a route in advance, it can be known post-hoc after observation. This leads us to the concept of ``uncertainty'' within the context of a dynamic patrolling scenario. Under the assumptions that edge weights vary according to an unknown time-dependent model and can only be observed by traversing the edge, any edge which has not been traversed in a long time (and so may have varied significantly since it was last observed) can be considered to have high uncertainty, and any edge which has been recently traversed has a low uncertainty.

\section{Methods of Handling Dynamic Environments}

\subsection{Key Principles and Constraints}

Two key principles guided the approaches taken in this work. Firstly, that the goal is to improve idleness-minimising behaviour in the presence of environment dynamics, rather than to develop the best model or understanding of the environment. This is a subtle line to walk --- literature interest in exploration of dynamic or uncertain environments often considers the understanding of the environment to be the goal. We, in contrast, already have an established goal in this case --- we cannot compromise idleness minimisation in order to collect information, unless we are confident that information would lead to improved idleness minimisation performance. Our second key principle is that any method developed should operate seamlessly alongside existing idleness-minimising decentralised multi-agent patrol strategies, in order to be deployable without requiring modification of patrol strategies. To this end, our method cannot directly interfere with decision making, as this is entirely controlled by the patrol strategy, and must therefore influence agent behaviour by modifying the inputs to the patrol strategy --- specifically, observations and beliefs in the state of the environment. 

We constrain ourselves further by noting that a desirable method to address this problem should be model-free. Any method that involves building an internal model of environment dynamics has the flaw that, in general, there is no reason to assume that the environment dynamics are controlled or modellable by a persistent underlying model. The nature of the dynamics may change rapidly and unpredictably, resulting in a previously fit model now being entirely inappropriate. The lack of a model limits our ability to predict future behaviors, steering us towards a purely reactive approach. This also precludes the use of a Bayesian or Kalman filter-based approach, both of which are otherwise obvious choices for modelling uncertain scenarios.

\subsection{The decay method}

To satisfy these principles, we use the \textit{decay} method, a case of the three-parameter RE (Roth/Erev) method~\cite{erev:RL}, a classical game-theoretical model-free reinforcement learning approach. This method has seen attention in the literature as a decision making method in uncertain environments~\cite{speekenbrink:bandits}\cite{ahn:decisions}, and our naming of it follows these works. Here we show that our particular problem is a natural fit for a special case of this method. Following~\cite{erev:RL}, the three-parameter RE method for decision making in uncertain environments is as follows:
\vspace{-0.5mm}
\begin{equation}
    q_{nj}(t+1) = (1-\phi)q_{nj}(t) + E_k(j, R(x))
\end{equation}

Where $q_{nj}(t)$ is the propensity for player $n$ to play strategy $j$ at time $t$, $\phi$ is a recency parameter to gradually reduce the importance of past experience, and $E(j, R(x))$ is an update to each strategy $j$ according to the experience of playing strategy $k$ and receiving reward $R(x)$. Fitting this method to our scenario affords us a natural approach to handling the tradeoff between exploration (observing uncertain edge weights) and exploitation (using knowledge to minimise idleness).

First, we argue that the propensity to play a strategy can be substituted for the belief of an agent in the weight of an edge of the patrol graph. A rational idleness-minimising agent will, all other things being equal, preferentially traverse a lower-weight edge than a higher-weight edge in order to sooner reduce the idleness of a vertex. This then allows us to encourage or discourage certain actions by modifying an agent's belief in the weight of an edge --- for example, reducing the weight belief on a certain edge will encourage an agent to traverse it at a time when doing so will still appear to be rational idleness minimising behaviour. In this way it is possible to encourage exploration without requiring any modification of an idleness-minimising patrol strategy, and avoid forcing exploratory actions without considering idleness minimisation.

We must then select $E_k(j, R(x))$ such that upon traversing an edge $j$, $q_j$ is set to the observed edge weight (as the agent now has high certainty in the edge weight) and all other edge weights are updated to reflect the increase in uncertainty caused by time passing without traversal. In this problem, the best representation of ``higher uncertainty'' in an edge weight is that in decision making its value is moved closer to the mean average current edge weight belief over the entire graph, as this represents the ``zero information'' state when comparing edges to each other. This ensures that, all other things being equal, a high uncertainty edge will be considered more desirable than one that is known to have a larger (worse) than average weight, and less desirable than one that is known to have a smaller (better) than average weight. This formulation of the three-parameter RE method has previously been seen in the literature~\cite{speekenbrink:bandits}\cite{ahn:decisions} within the context that in a scenario in which rewards are symmetrically distributed around zero, expected values of actions decay towards zero (i.e. no information) the longer they are unchosen.

Our implementation of this, which we refer to as the \textit{decay} method, is as follows. When an agent finishes traversing an edge, it records the time taken as the weight of that edge. This is transmitted to other agents, who do the same upon receipt of the message. The agent's beliefs of all other edge weights then decay towards the collective mean $\overline{W}$ based on a decay factor $\phi$, such that:
\vspace{-0.5mm}
\begin{equation}
    W_i = (1-\phi)^t W_{i-1} + \phi^t \overline{W}
\end{equation}

Where $t$ is the time since the agent last updated its belief in (or received new information of) a given edge weight. These weight beliefs are input to the patrol strategy being used by the patrol team. Our final addition to this method is a time-dependent decay factor to account for varying time between observations.  In this work, $\phi = 0.0025$ was used, tuned on two maps not used in our testing.

\subsection{Baseline methods}

\noindent Alongside our decay method, the following baseline methods are considered:

\begin{itemize}
    \item \textbf{Lazy}: The patrol agents do not monitor the environment, and make all decisions on the basis of their prior belief of environment edge weights.
    \item \textbf{Simple}: When a patrol agent finishes traversing an edge, it records the time taken as the weight of that edge. This is communicated to other agents, who do the same. Edge weight beliefs only change when the edge is traversed.
    \item \textbf{Omniscient}: The patrol agents have perfect knowledge of the environment dynamics and edge weights at all times.
\end{itemize}

\vspace{-2mm}
\section{Testing}
\vspace{-2mm}
\subsection{Constructed scenarios}
\label{sec:constructed}
To examine the performance of our decay method, we selected three leading literature patrol strategies to handle idleness-minimising decision making. State Exchange Bayesian Strategy (SEBS)~\cite{portugal:sebs_paper}, Expected Reactive (ER)~\cite{yan:er_paper}, and Minimal Network Strategy (MNS)~\cite{ward:suns_mns} were selected as all three are fully online decentralised strategies offering high levels of idleness minimisation performance. Also, all three have sufficiently different mechanisms, such that we are combining the dynamic environment handling methods with a range of idleness minimising behaviours. The four dynamic environment handling methods (``lazy'', ``simple'', ``omniscient'', and ``decay'') and three patrol strategies were implemented in MAGESim\footnote{\url{https://github.com/jward0/magesim}. Our implementations can be found on the ``dynamic\_environment'' branch.}, a simulator used for examining multi-agent behaviours in graph structured environments. 

\begin{figure}[H]
\vspace{-7mm}
\begin{minipage}{0.475\textwidth}
    \caption{The two patrol graphs used in our constructed scenarios (not to scale)}
    \label{fig:patrol_graphs}
    \includegraphics[width=\linewidth]{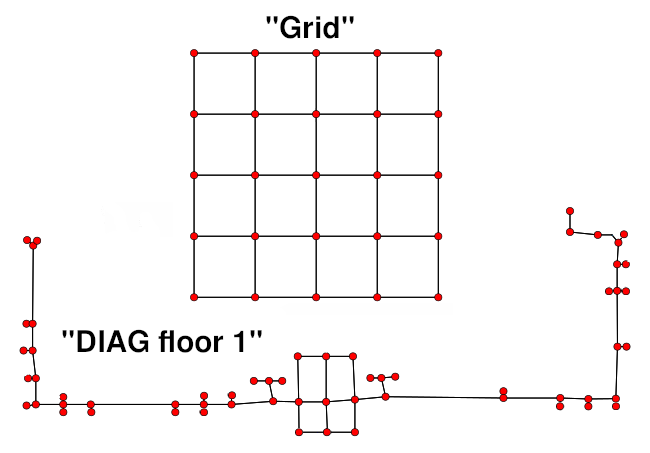}
\end{minipage}
\hfill
\begin{minipage}{0.475\textwidth}
    \caption{Example ``blockages'' dynamics}
    \label{fig:blockages}
    \includegraphics[width=\linewidth]{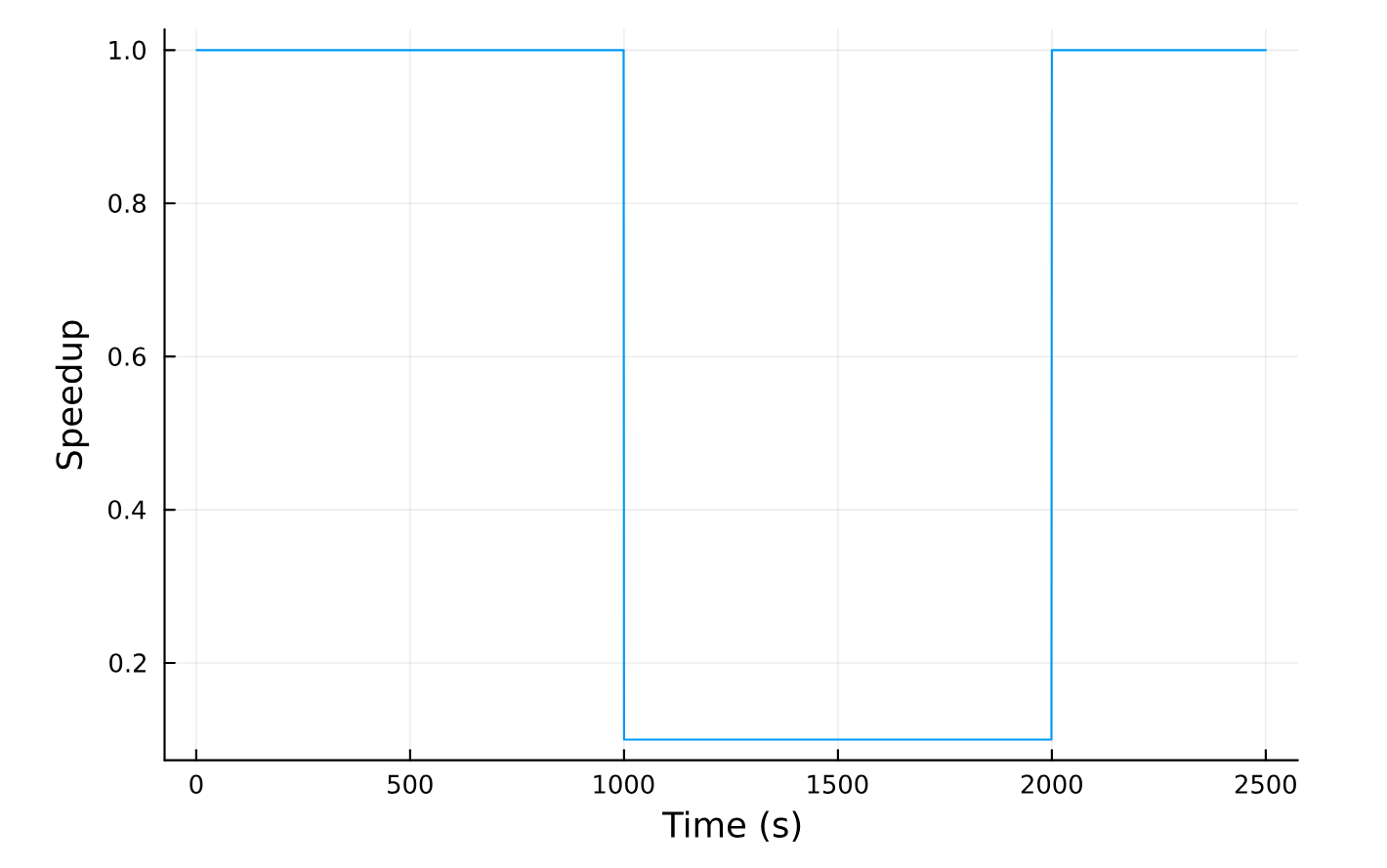}
\end{minipage}
\end{figure}

\begin{figure}[H]
\begin{minipage}{0.475\textwidth}
    \caption{Example ``fast walk'' dynamics}
    \label{fig:fast_walk}
    \includegraphics[width=\linewidth]{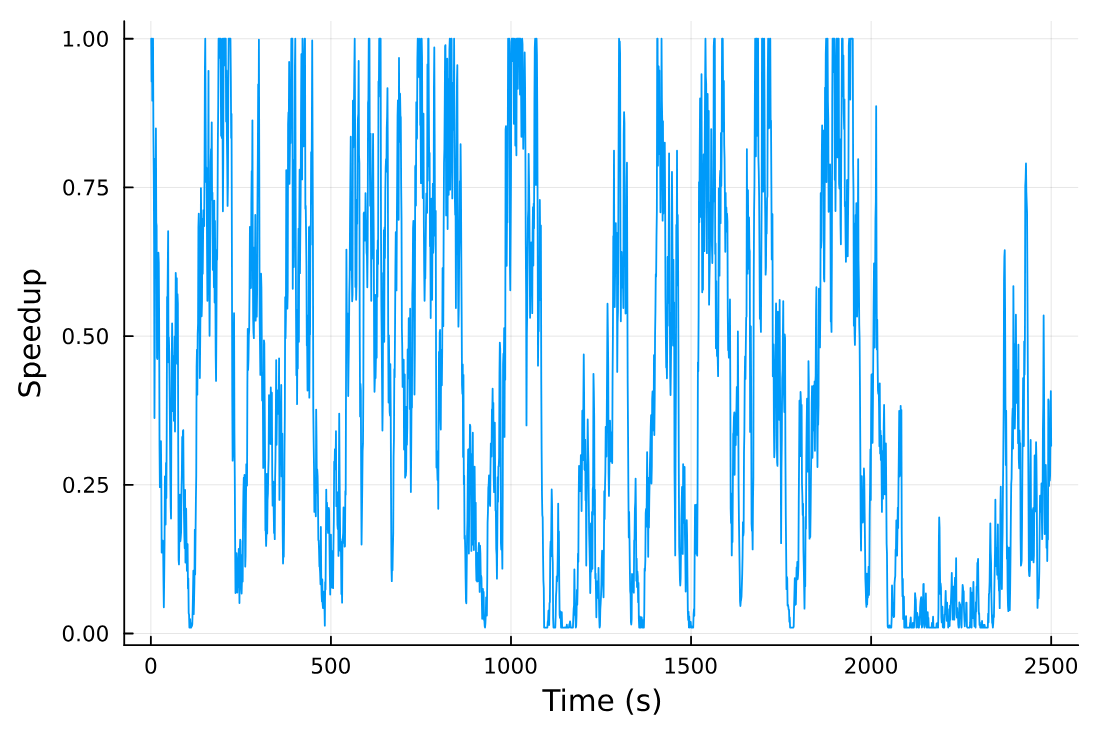}
\end{minipage}
\hfill
\begin{minipage}{0.475\textwidth}
    \caption{Example ``smooth walk'' dynamics}
    \label{fig:slow_walk}
    \includegraphics[width=\linewidth]{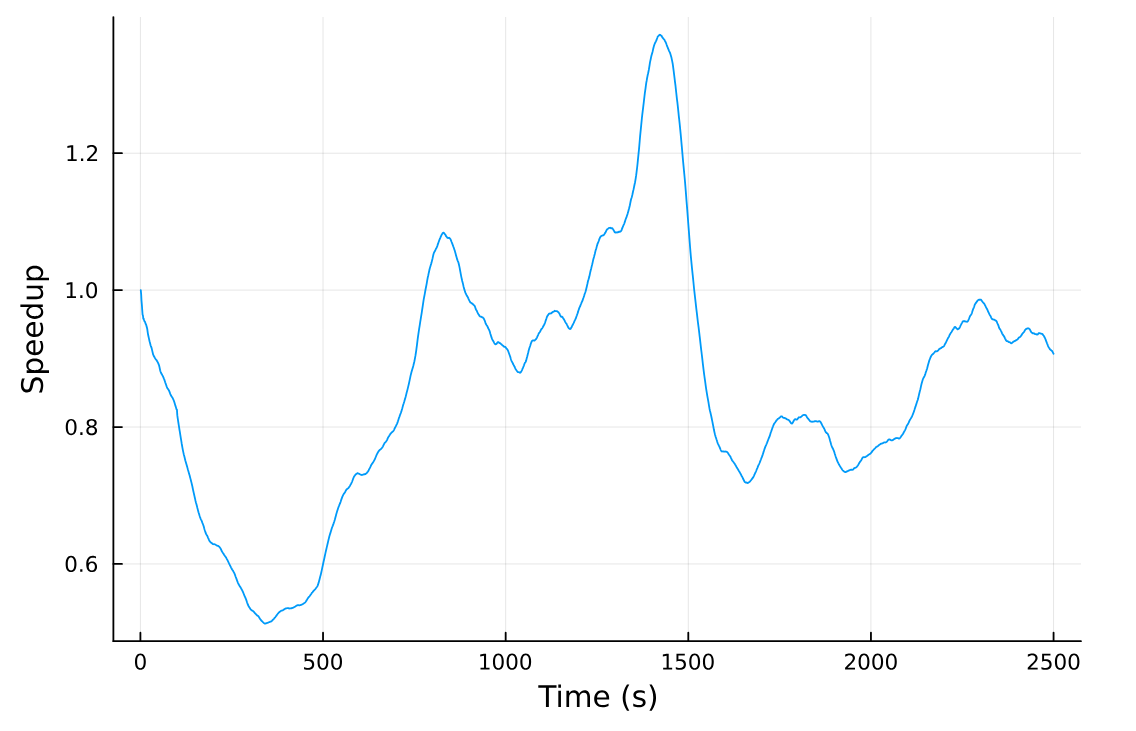}
\end{minipage}
\vspace{-5mm}
\end{figure}

Two maps, ``Grid'' and ``DIAG floor 1'' (originally from ROS Patrolling Sim\footnote{\url{http://wiki.ros.org/patrolling\_sim}}), shown in Figure~\ref{fig:patrol_graphs}, were selected. To model a range of environment dynamics, we generated three sets of dynamic profiles to apply to the edge weights of the patrol graphs. These operate by scaling the speed at which agents are able to move along an edge by a time-varying factor, to reflect edges becoming more or less easy to traverse at different times. The first, ``blockages'', applies a speed scale factor of 0.1 for long, randomly selected periods of time. The second, ``fast walk'', scales agent speed according to a fast-moving noisy random walk clamped to the range (0, 1]. The third, ``smooth walk'', scales agent speed according to a slower-moving random walk with momentum with a rolling average applied, clamped to all positive values (the smooth walk is then the only profile that can \textit{increase} agent speed along an edge). Examples of these profiles are shown in Figures~\ref{fig:blockages}--\ref{fig:slow_walk}. When one of these profiles is generated for a patrol graph, the speedup for each edge is generated randomly and independently, such that the average correlation between speedups on different edges is close to 0.

Patrol teams of size $n$ = 1, 2, 4, 8, and 16 were simulated for all combinations of map, dynamic profile, dynamic handling method, and patrol strategy, for 2500s. The prior edge weight beliefs for the lazy handling method were set to the edge weights in the absence of dynamics. For each combination of parameters, the simulation was repeated 5 times with different initial agent positions. We additionally simulated all scenarios with no environment dynamics present. 

To assess the robustness of the patrol strategies to incorrect information, these tests were repeated with no dynamics present, the lazy dynamic handling method, and multiplicative Gaussian noise ($\mu=1$, $\sigma=0.05$ to $1.60$) applied to the patrol team's beliefs in the edge weights. This allowed us to measure performance in the cases where the patrol teams were making decisions based on consistently incorrect information.

\vspace{-15mm}

\subsection{Real-world scenario}
\label{section:real-world}

\begin{figure}[H]
\begin{minipage}{0.4\textwidth}
    \includegraphics[width=\linewidth]{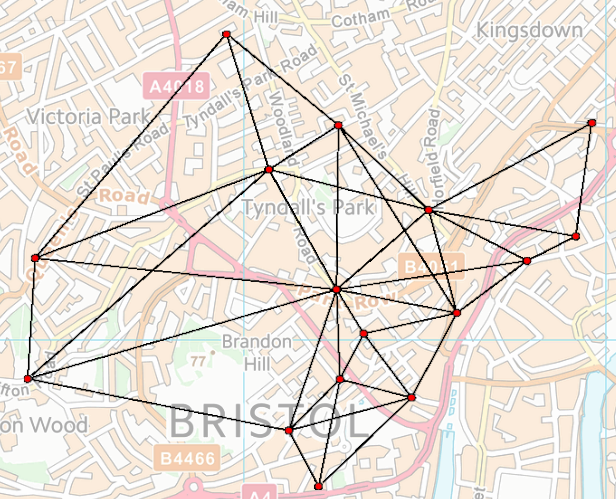}
    \vspace{-2mm}
    \vspace{1mm}
    \caption{Patrol graph used in the traffic scenario\protect\footnotemark}
    \label{fig:traffic_graph}
\end{minipage}
\hfill
\begin{minipage}{0.55\textwidth}
    \includegraphics[width=\linewidth]{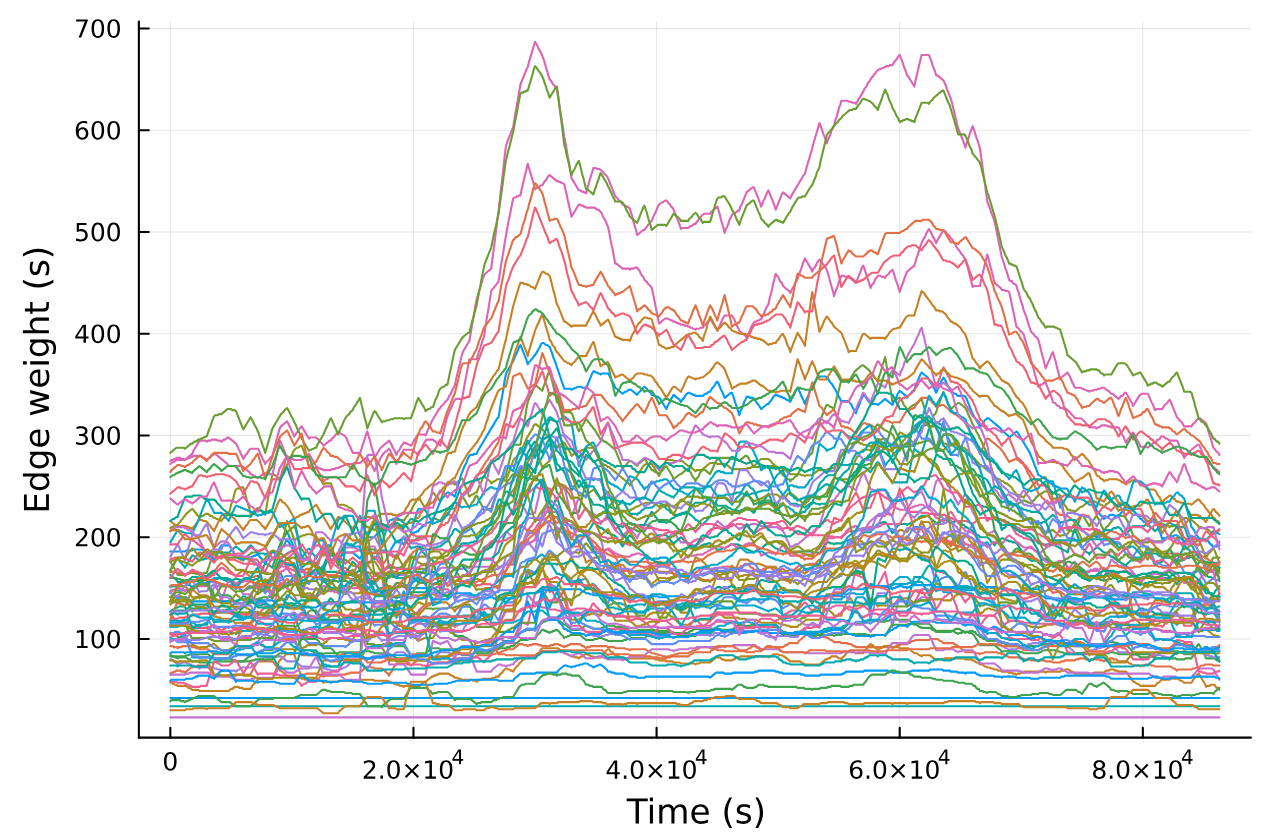}
    \vspace{-6mm}
    \caption{Edge weights over time for the traffic scenario (each line represents one edge)}
    \label{fig:traffic_edges}
\end{minipage}
\end{figure}

In order to examine the behaviour of patrol teams in a real-world dynamic environment, we constructed a scenario based on real-world traffic data. 16 points were selected throughout the city of Bristol from locations of University accommodation and used to construct a patrol graph, shown in Figure~\ref{fig:traffic_graph}. Edge weights corresponding to vertex-to-vertex travel times were then obtained from predicted traffic data over a 24 hour period starting at midnight, using the Google Maps Routes API to estimate travel times along all edges at 10 minute intervals, following the fastest predicted routes at those times. These edge weights are shown in Figure~\ref{fig:traffic_edges}. These weights were then converted to a dynamic profile and imported into MAGESim, to allow for simulation based on real-world traffic data. As before, we selected SEBS, ER, and MNS as our patrol strategies, and examined the lazy, simple, omniscient, and decay dynamic environment handling methods. We simulated patrol teams of size $n$ = 1, 2, 4, and 8 (16 agents on a 16-vertex graph was deemed unrealistically crowded) for the full 24 hour (86400s) period. As before, we also simulated this environment with no dynamics (generated by artificially setting all edge weights to their smallest observed values to reflect travel times with no congestion). It is worth noting that in this scenario, the adjacency matrix of the patrol graph is not symmetric, meaning that traversal times along an edge vary depending on the direction of travel --- this was not the case in our constructed scenarios (Section \ref{sec:constructed}), but is obviously possible in a real-world scenario.

\vspace{-2mm}
\vspace{-7mm}
\footnotetext{Map image provided by Ordnance Survey under the Open Government License (\url{https://www.nationalarchives.gov.uk/doc/open-government-licence/version/3/})}
While our constructed scenarios involved generating dynamic profiles for all edges randomly and independently, edge weights in this scenario are strongly correlated, with average correlation coefficient between any two edges of 0.736. This had a significant impact on behaviour, as we will discuss in the next sections.
\vspace{-8mm}
\section{Results}
\vspace{-1mm}
\label{sec:results}
\subsection{Constructed scenarios}
Table~\ref{table:baseline} shows the relative average idlenesses (measured average idlenesses divided by those observed with no environment dynamics) averaged across both maps for all scenarios considered for our three dynamic profiles, using the lazy baseline method. This is a measure of the impact of the environment dynamics in the case that nothing is done to account for them, to quantify the degree to which they interfere with idleness minimising behaviour. This shows that fast walk is the most disruptive profile, followed by blockages and then smooth walk, and also that relative idleness generally increases with the number of agents --- we attribute this to the fact that as the number of agents increases so does the importance of effective inter-agent coordination, which is compromised by having incorrect environmental information. In the case where no environment dynamics were present, none of the dynamic handling methods tested were observed to significantly alter performance. 
\vspace{-2mm}
\begin{table}[H]
    \vspace{-2mm}
    \centering
    \caption{Average idlenesses relative to no environmental dynamics for lazy baseline, varying with team size $n$}
    \vspace{-2mm}
    \label{table:baseline}
    \begin{tabular}{cccccccccccccc}
        ~ & \multicolumn{3}{c}{Blockages} & ~ & ~ & \multicolumn{3}{c}{Fast walk} & ~ & ~ & \multicolumn{3}{c}{Smooth walk} \\
        \cline{1-4} \cline{7-9} \cline{12-14}
        $n$ & SEBS & ER & MNS & ~ & ~ & SEBS & ER & MNS & ~ & ~ & SEBS & ER & MNS \\ 
        \cline{1-4} \cline{7-9} \cline{12-14}
        1 & 3.10 & 2.54 & 2.49 & ~ & ~ & 4.14 & 3.71 & 3.65 & ~ & ~ & 1.19 & 1.18 & 1.13 \\ 
        2 & 3.01 & 2.67 & 2.84 & ~ & ~ & 4.66 & 5.12 & 5.07 & ~ & ~ & 1.20 & 1.22 & 1.19 \\ 
        4 & 3.06 & 3.12 & 2.97 & ~ & ~ & 5.77 & 5.38 & 5.47 & ~ & ~ & 1.22 & 1.22 & 1.26 \\ 
        8 & 3.40 & 3.76 & 3.12 & ~ & ~ & 5.94 & 6.28 & 5.86 & ~ & ~ & 1.30 & 1.37 & 1.28 \\ 
        16 & 3.37 & 4.26 & 2.98 & ~ & ~ & 7.20 & 7.70 & 6.70 & ~ & ~ & 1.27 & 1.45 & 1.26 \\ 
        \cline{1-4} \cline{7-9} \cline{12-14}
    \end{tabular}
    \vspace{-4mm}
\end{table}

\subsubsection{Dynamic handling}
~ \newline
The results in this section consist of average idlenesses normalised against those observed using the aforementioned lazy baseline in identical scenarios. This allows us to directly examine the degree to which the methods considered present an improvement over default patrol behaviour that makes no effort to account for varying environments.

Tables~\ref{table:profile_varying} and~\ref{table:teamsize_varying} show relative average idlenesses observed with the simple and omniscient baselines and our decay method, varying with dynamic profile and team size. Relative idlenesses averaged over every variable for the three methods are $0.934$ for the simple method, $0.920$ for the decay method, and $0.846$ for the omniscient method, representing improvements over the lazy baseline of $6.6\%$, $8.0\%$, and $15.4\%$ respectively. The full un-averaged results were examined using the Wilcoxon signed-rank test, as this allows us to match results from identical scenarios for each method. This found that while the omniscient method unsurprisingly outperforms the others, our decay method outperforms the simple method with $p$-value of $0.038$. While statistically significant to within $p = 0.05$, the effect size ($8.0\%$ versus $6.6\%$ reduction in average idleness) is not large. We discuss this further in Section~\ref{sec:discussion}.

\begin{table}[h]
    \centering
    \caption{Average idlenesses relative to lazy baseline varying with dynamic profile}
    \vspace{-2mm}
    \label{table:profile_varying}
    \begin{tabular}{cccccccccccccc}
        ~ & \multicolumn{3}{c}{Simple} & ~ & ~ & \multicolumn{3}{c}{Decay} & ~ & ~ &\multicolumn{3}{c}{Omniscient} \\
        \cline{1-4} \cline{7-9} \cline{12-14}
        Profile & SEBS & ER & MNS & ~ & ~ & SEBS & ER & MNS & ~ & ~ & SEBS & ER & MNS \\ 
        \cline{1-4} \cline{7-9} \cline{12-14}
        Blockages & 0.78 & 0.79 & 0.97 & ~ & ~ & 0.77 & 0.79 & 0.96 & ~ & ~ & 0.68 & 0.71 & 0.93 \\ 
        Fast walk & 1.05 & 1.09 & 1.01 & ~ & ~ & 0.99 & 1.02 & 1.01 & ~ & ~ & 0.82 & 0.86 & 0.95 \\ 
        Smooth walk & 0.89 & 0.86 & 0.96 & ~ & ~ & 0.89 & 0.87 & 0.97 & ~ & ~ & 0.86 & 0.84 & 0.96 \\ 
        \cline{1-4} \cline{7-9} \cline{12-14}
    \end{tabular}
    \vspace{-1mm}
\end{table}

\begin{table}[h]
    \vspace{-0mm}
    \centering
    \caption{Average idlenesses relative to lazy baseline varying with team size $n$}
    \vspace{-2mm}
    \label{table:teamsize_varying}
    \begin{tabular}{cccccccccccccc}
        ~ & \multicolumn{3}{c}{Simple} & ~ & ~ & \multicolumn{3}{c}{Decay} & ~ & ~ & \multicolumn{3}{c}{Omniscient} \\
        \cline{1-4} \cline{7-9} \cline{12-14}
        $n$ & SEBS & ER & MNS & ~ & ~ & SEBS & ER & MNS & ~ & ~ & SEBS & ER & MNS \\ 
        \cline{1-4} \cline{7-9} \cline{12-14}
        1 & 1.01 & 1.01 & 0.98 & ~ & ~ & 0.95 & 0.97 & 0.99 & ~ & ~ & 0.87 & 0.91 & 0.92 \\ 
        2 & 0.99 & 1.00 & 0.98 & ~ & ~ & 0.94 & 0.96 & 1.01 & ~ & ~ & 0.86 & 0.90 & 0.93 \\ 
        4 & 0.93 & 0.97 & 1.01 & ~ & ~ & 0.90 & 0.94 & 0.99 & ~ & ~ & 0.81 & 0.83 & 0.98 \\ 
        8 & 0.83 & 0.83 & 0.98 & ~ & ~ & 0.84 & 0.82 & 0.98 & ~ & ~ & 0.71 & 0.71 & 0.97 \\ 
        16 & 0.79 & 0.76 & 0.95 & ~ & ~ & 0.79 & 0.76 & 0.94 & ~ & ~ & 0.71 & 0.66 & 0.93 \\ 
        \cline{1-4} \cline{7-9} \cline{12-14}
    \end{tabular}
    \vspace{-6mm}
\end{table}

\subsubsection{Robustness to incorrect information} 
~ \newline
Table~\ref{table:noise} shows average relative idleness against the standard deviation of the multiplicative 1-mean Gaussian noise applied to the patrol team's belief of edge weights with no environment dynamics. The performance of SEBS and ER was observed to not significantly worsen until $\sigma = 0.4$ (corresponding to an average error magnitude of roughly 32\%) after which performance gradually decayed at a similar rate for both, while MNS was observed to have extremely effective noise-rejection behaviour, as no degradation in performance was observed even with the largest noise magnitude tested. This high level of apparent robustness is discussed in Section~\ref{sec:discussion}.

\begin{wraptable}{r}{0.45\textwidth}
\vspace{-28mm}
    \centering
    \caption{Average relative idleness against noise standard deviation (constructed scenarios)}
     \label{table:noise}
     \vspace{1mm}
     \begin{tabular}{cccc}
     \hline
         $\sigma$ & SEBS & ER & MNS \\ \hline
         0.05 & 1.00 & 0.99 & 1.00 \\ 
         0.10 & 1.00 & 0.99 & 0.99 \\ 
         0.20 & 1.00 & 0.99 & 0.99 \\ 
         0.40 & 1.06 & 1.04 & 0.98 \\ 
         0.80 & 1.09 & 1.10 & 0.99 \\ 
         1.60 & 1.14 & 1.16 & 0.99 \\ 
         \hline
     \end{tabular}

         \caption{Average idlenesses relative to no dynamics for lazy baseline, varying with team size $n$ (real-world scenario)}
         \vspace{1mm}
         \label{table:real_baseline}
         \begin{tabular}{cccc}
             \hline
             $n$ & SEBS & ER & MNS\\ 
             \hline
             1 & 1.33 & 1.33 & 1.35 \\ 
             2 & 1.24 & 1.27 & 1.30 \\ 
             4 & 1.27 & 1.31 & 1.29 \\ 
             8 & 1.29 & 1.28 & 1.31 \\ 
             \hline
         \end{tabular}
         \vspace{-4mm}
\end{wraptable}

\subsection{Real-world scenario}

Table~\ref{table:real_baseline} shows the relative average idlenesses of the observed real-world traffic scenario compared to the same environment with congestion artificially removed (i.e. edge weights are constant at their smallest observed values) in order to quantify the impact of the environment dynamics. Table~\ref{table:realworld} shows the relative average idlenesses observed with the simple, decay, and omniscient methods compared to the lazy baseline for all patrol strategies and team sizes tested.  While the observed average slowdown from the environment dynamics was larger than was observed for smooth walk in our constructed scenarios, omniscient handling only offered an average improvement of $1.8\%$ over lazy handling, despite having access to perfect information.  We also observe that neither the simple baseline nor our decay method offer improved performance over the lazy baseline. This suggests that, in this real-world scenario, improved access to information does not significantly improve performance. We discuss this phenomenon in Section~\ref{sec:discussion}. As with our constructed scenarios, none of the dynamic handling methods were observed to significantly alter performance in the absence of environment dynamics.

\begin{table}[H]
    \centering
    \vspace{-3mm}
    \caption{Average idlenesses relative to lazy baseline varying with team size $n$}
    \vspace{-2mm}
    \label{table:realworld}
    \begin{tabular}{cccccccccccccc}
        ~ & \multicolumn{3}{c}{Simple} & ~ & ~ & \multicolumn{3}{c}{Decay} & ~ & ~ & \multicolumn{3}{c}{Omniscient} \\
        \cline{1-4} \cline{7-9} \cline{12-14}
        $n$ & SEBS & ER & MNS & ~ & ~ & SEBS & ER & MNS & ~ & ~ & SEBS & ER & MNS \\ 
        \cline{1-4} \cline{7-9} \cline{12-14}
        1 & 1.00 & 1.00 & 1.05 & ~ & ~ & 0.94 & 0.94 & 1.01 & ~ & ~ & 0.94 & 0.94 & 1.01 \\ 
        2 & 1.20 & 1.18 & 1.05 & ~ & ~ & 1.04 & 1.01 & 1.00 & ~ & ~ & 1.01 & 0.99 & 1.01 \\ 
        4 & 1.18 & 1.13 & 1.05 & ~ & ~ & 1.07 & 1.00 & 1.01 & ~ & ~ & 0.99 & 0.96 & 1.00 \\ 
        8 & 1.06 & 1.14 & 1.01 & ~ & ~ & 0.97 & 1.04 & 1.00 & ~ & ~ & 0.98 & 0.96 & 0.99 \\ 
        \cline{1-4} \cline{7-9} \cline{12-14}
    \end{tabular}
    \vspace{-8mm}
\end{table}

\vspace{-3mm}
\section{Discussion}
\label{sec:discussion}
\vspace{-3mm}
We have presented a practical and effective method of handling environments with time-varying traversability in decentralised on-line multi-robot patrol. Our results from our proposed decay method show an average idleness reduction of $8\%$ compared to the lazy baseline over our constructed scenarios. While the omniscient baseline offered better performance, it is impractical to implement in a real patrol scenario. The simple baseline performed only slightly (but statistically significantly) worse than the decay method, with average idleness reduction of $6.6\%$ compared to the lazy baseline. We attribute the small difference of effect size to the similarity between our decay method and the simple baseline combined with the robustness of the patrol strategies to incorrect information. Our examination of this, by adding multiplicative Gaussian noise to the edge weight belief of patrol teams, demonstrates that the patrol strategies used are highly robust to incorrect patrol weight belief, with no observed drop in performance for SEBS or ER until $\sigma = 0.40$, corresponding to an average error magnitude of roughly 32\%. As such, the disregard for information staleness by the simple method could be largely absorbed by the robustness of the patrol strategies.

MNS, meanwhile, demonstrated no performance degradation even for the largest noise magnitudes examined --- this apparent rejection of edge weight information was reflected in our dynamic environment handling results, for which all methods show the least improvement for MNS. To investigate this, we examined the neural network weights used in the pre-trained, published instance of MNS that was used in this work, finding that MNS assigns only very small weights to edge information. This is counter-intuitive, as it suggests that leading levels of idleness minimisation are attainable while paying very little attention to the actual structure of the patrol graph, and instead making decisions purely on the basis of vertex idlenesses. Its apparent robustness to incorrect information was also observed in the average idlenesses relative to no dynamics for lazy baseline, shown in Table~\ref{table:baseline} --- MNS suffers a smaller degradation in performance as a result of unaccounted environment dynamics than either SEBS or ER.

A second observation from our constructed scenarios is that the fast walk dynamics proved challenging for non-omniscient methods to deal with (though our decay method did still outperform the simple baseline). This is unsurprising, as these dynamics vary sufficiently quickly (see Figure~\ref{fig:fast_walk}) that any observed edge weights are invalidated almost immediately. In this case, then, ignoring the environment dynamics altogether may be a reasonable approach --- it may also be true that the fast walk dynamic model presents an unrealistically harsh scenario that would be unlikely to appear in a real deployment.

In our simulation of a real-world traffic scenario, the omniscient baseline was found to outperform the lazy baseline by only a tiny margin, and neither of the other methods offered improvements. We attribute this to the strong inter-correlation ($\overline{\rho} = 0.736$) between observed edge weights over time. In Figure~\ref{fig:traffic_edges}, this can be observed as all edge weights following similar patterns of variation, just at different scales. The consequence of this is that the relative attractiveness of the different edges varied very little over time --- edges that were attractive for low congestion were still attractive for high congestion, so decisions can be made effectively on the basis of a snapshot of the edge weights without time-varying information. This has important implications for assessing when environment dynamics in a patrol scenario can be safely ignored --- even if there is significant variation in edge weights over time, if these variations for all edges are strongly correlated with each other then none of the methods discussed in this paper are necessary to achieve high performance.

Based on our results, we suggest that, in a potential real-world deployment, plausible models of environmental dynamics should be considered to see whether an explicit dynamic handling approach is useful. If dynamics are too highly variable, as in our fast walk profile, or too highly inter-correlated, as in our real-world traffic scenario, on-line adaptation may simply not be an effective or necessary tool to improve performance. However, in cases where dynamics vary on a timescale that observations made by one agent can be useful to another and are not strongly inter-correlated, as was observed in our blockages and smooth walk profiles, the model-free decay method presented in this work is a practical and effective on-line, decentralised approach to maintaining high levels of performance, when used in tandem with an existing patrol strategy. 
\vspace{-3mm}
\section{Conclusions}
\vspace{-3mm}
In this work, we have examined how dynamic environments might be practically handled in a decentralised, fully on-line multi-robot patrol deployment. Considering several models of environment dynamics, we have identified scenarios in which intelligent handling of said dynamics are either unnecessary or impractical. In cases where the environment dynamics can be effectively addressed with a practically implementable method, we have shown that our proposed ``decay'' method outperforms plausible baselines. This method can integrate with existing idleness-minimising patrol strategies with no modification required, can operate in a fully decentralised, on-line fashion, requires no additional observation or inter-agent communication beyond that already carried out during patrolling, and does not compromise performance in dynamically inactive environments. As such, our proposed method presents a practical, effective approach to improving performance in potential real-world deployments.

\begin{credits}
\subsubsection{\ackname} This work was supported by the Engineering and Physical Sciences Research Council via the FARSCOPE-TU CDT (grant no. EP/S021795/1) and by the Royal Academy of Engineering under the Research Fellowship programme.

\subsubsection{\discintname}
The authors have no competing interests to declare that are
relevant to the content of this article.
\end{credits}

\newpage
\bibliographystyle{splncs04}
\bibliography{bib}

\end{document}